\documentclass[11pt]{article}

\usepackage[margin=1in]{geometry}
\usepackage[T1]{fontenc}
\usepackage[utf8]{inputenc}
\usepackage{mathptmx}      
\usepackage{microtype}
\usepackage[hidelinks]{hyperref}
\usepackage{xcolor}

\usepackage{amsmath}
\usepackage{amssymb}
\usepackage{algorithm}
\usepackage{algpseudocode}
\usepackage{multirow}
\usepackage{subfigure}     
\usepackage{graphicx}

\title{}   
\author{}  
\date{}    

\begin{document}

\begin{center}
    \noindent\rule{0.92\textwidth}{1.8pt}\par
    \vspace{1.0em}

    \begin{minipage}{0.92\textwidth}
        \centering
        {\LARGE \bfseries
        Conformalized Transfer Learning for Li-ion Battery State of Health Forecasting under Manufacturing and Usage Variability\par}
    \end{minipage}

    \vspace{1.0em}
    \noindent\rule{0.92\textwidth}{1.8pt}\par
    \vspace{0.9em}

    {\large \scshape A Preprint\par}

    \vspace{1.0em}
    Samuel Filgueira da Silva$^{1}$, Mehmet Fatih Ozkan$^{1}$, Faissal El Idrissi$^{1}$, and Marcello Canova$^{1}$\par

    \vspace{0.4em}
    {\small
    $^{1}$Department of Mechanical and Aerospace Engineering, Center for Automotive Research,\\
    The Ohio State University, Columbus, OH, USA\par}

    \vspace{0.25em}
    {\small
    \texttt{filgueiradasilva.1@osu.edu, ozkan.25@osu.edu, elidrissi.2@osu.edu, canova.1@osu.edu}\par}

    \vspace{0.4em}
    {\small \today\par}
\end{center}

\vspace{1em}

\begin{abstract}
Accurate forecasting of state-of-health (SOH) is essential for ensuring safe and reliable operation of lithium-ion cells. However, existing models calibrated on laboratory tests at specific conditions often fail to generalize to new cells that differ due to small manufacturing variations or operate under different conditions. To address this challenge, an uncertainty-aware transfer learning framework is proposed, combining a Long Short-Term Memory (LSTM) model with domain adaptation via Maximum Mean Discrepancy (MMD) and uncertainty quantification through Conformal Prediction (CP). The LSTM model is trained on a virtual battery dataset designed to capture real-world variability in electrode manufacturing and operating conditions. MMD aligns latent feature distributions between simulated and target domains to mitigate domain shift, while CP provides calibrated, distribution-free prediction intervals. This framework improves both the generalization and trustworthiness of SOH forecasts across heterogeneous cells.
\end{abstract}

\section{Introduction}

Lithium-ion batteries play a dominant role in the transition toward electrified transportation and renewable energy integration, although their limited lifespan remains a critical technological bottleneck. Accurate prediction of the battery state of health (SOH) is essential not only for ensuring safety and reliability, but also for enabling advanced energy management strategies such as aging-aware fast charging \cite{lu2024health}. Despite their importance, capturing SOH trajectories over long horizons remains challenging due to the complex, nonlinear, and often coupled nature of degradation mechanisms \cite{edge2021lithium}. 

Physics-based models, such as the Pseudo-Two-Dimensional (P2D) and Extended Single Particle Model (ESPM), provide valuable insights into electrochemical processes and can be augmented with degradation submodels such as solid electrolyte interphase (SEI) growth, lithium plating, and loss of active material \cite{birkl2017degradation}. These models have the advantage of interpretability and can simulate aging under diverse operating conditions \cite{marcicki2013design}. 

Recent studies have also shown that deep learning (DL) models can effectively uncover nonlinear mappings between usage patterns and degradation outcomes \cite{LI2019510}. However, data-driven models face challenges in terms of robustness when applied to previously unseen conditions. One of the critical challenges arises from cell-to-cell variability \cite{schindler2021evolution}, where even minor deviations in electrode manufacturing can yield significant changes in degradation behavior \cite{weng2023differential}. Traditional SOH forecasting models are generally trained on nominal cells, limiting their ability to generalize to samples that deviate from the average population. Consequently, there is a need for modeling approaches that can effectively leverage knowledge from a generic population of cells while transferring it to non-nominal cases.

Transfer learning (TL) has emerged as a strategy to address this issue. By transferring knowledge from a well-characterized source domain to a less-represented target domain, TL enables robust forecasting under distribution shifts, such as variations in temperature, cycling rates, and cell configurations. Recent works have employed TL-assisted SOH forecasting, using fine-tuning \cite{shu2021flexible}, metric-based domain adaptation techniques \cite{han2022end, ma2022transfer} based on Maximum Mean Discrepancy (MMD) and Correlation Alignment (CORAL), and adversarial approaches such as Domain-Adversarial Neural Networks (DANNs) \cite{ye2021state,da2020remaining}. These methods have demonstrated improved generalization across operating conditions, but limited attention has been given to TL across cell-to-cell variability caused by manufacturing processes.

Another key challenge is quantifying the reliability of predictions. Existing approaches to uncertainty quantification (UQ), including Bayesian deep learning \cite{li2020bayesian}, Gaussian processes \cite{liu2020data}, and Monte Carlo dropout \cite{li2024sensor}, often rely on distributional assumptions that lead to overconfident estimates under real-world variability.

This paper proposes an uncertainty-aware, data-driven framework for SOH forecasting that addresses variability induced by manufacturing processes and operating conditions. The approach integrates an LSTM model with domain-adaptive transfer learning to improve generalization, while incorporating Conformal Prediction (CP) to provide distribution-free, finite-sample uncertainty quantification. The key contributions of this work are the development of a transfer learning framework that enables SOH forecasting across cells with manufacturing and usage-induced variability, and the integration of CP to deliver calibrated uncertainty intervals under variable and data-scarce conditions.


\section{Overview of LiB Cell Model Equations} \label{section_2}


This study adopts the Single Particle Model with electrolyte (SPMe) implemented in PyBaMM. SPMe extends the traditional Single Particle Model by including electrolyte concentration and potential dynamics, improving voltage prediction under high C-rates. For brevity, the full set of governing equations is omitted here and can be found in \cite{marquis2019asymptotic}. The terminal voltage is expressed as:
\begin{equation} \label{eq:1}
\begin{split}
V_{\mathrm{SPMe}}(t) &= U_p(c_{s,p},t) - U_n(c_{s,n},t) - \big( \eta_p(0,t) - \eta_n(L,t) \big) \\
&\quad - \big( \phi_{e}(0,t) - \phi_{e}(L,t) \big) - I(t)R_c,
\end{split}
\end{equation}

\noindent where $U_i$ are the electrode open-circuit potentials, $\eta_i$ the kinetic overpotentials, $\phi_e$ the electrolyte potential, $c_{s,i}$ the solid-phase lithium concentration (with $i = n,p$), and $R_c$ the lumped contact resistance.

\subsection{Battery Degradation Model}
To capture long-term capacity fade and performance loss, the SPMe is augmented with degradation submodels for loss of active material (LAM) and loss of lithium inventory (LLI) driven by SEI growth. The adopted formulations are consistent with physics-based capacity-loss models for graphite anodes and SEI/LAM mechanisms reported in the literature \cite{birkl2017degradation,jin2017physically}. The model has previously been calibrated and validated against experimental capacity-fade datasets for an A123 26650 cell \cite{salyer2021extended}, supporting its use for SOH prediction.

\subsubsection{Loss of Active Material}
LAM accounts for the progressive deactivation of electrochemically active particles due to mechanical stress, electrode cracking, or reactions that isolate active material. It is modeled as a time-dependent decrease in the active material volume fraction $\varepsilon_{\mathrm{AM},i}$ of electrode $i \in \{n,p\}$:
\begin{equation}
  \frac{d \varepsilon_{\mathrm{AM},i}}{dt}
  = -\, k_{\mathrm{AM},i}\;
     \frac{3 R_i}{\varepsilon_{\mathrm{AM},i}\, A\, L_i}\;
     \bigl| I(t) \bigr|\;
     \exp\!\left(-\frac{E_{\mathrm{AM},i}}{RT}\right),
  \label{eq:lam}
\end{equation}
\noindent where $\varepsilon_{\mathrm{AM},i}$ is the active-material volume fraction, $R_i$ the particle radius, $L_i$ the electrode thickness, $A$ the electrode area, $I(t)$ the applied current, and $k_{\mathrm{AM},i}$ and $E_{\mathrm{AM},i}$ empirical parameters. Decreasing $\varepsilon_{\mathrm{AM},i}$ reduces the available cyclable capacity of the corresponding electrode.

\subsubsection{SEI Layer Growth}
Lithium loss due to the SEI layer growth is produced by the side reaction at the anode caused by the ethylene carbonate (EC) reduction in the organic electrolyte compounds. The formation of SEI consumes cyclable lithium, reducing the inventory available for intercalation. Following \cite{safari2008multimodal}, the side reaction current density is modeled using a cathodic Tafel expression:
\begin{equation}
i_{s} \;=\; 
-\,F \, k_{f,s} \, c_{\mathrm{EC}}
\exp\!\left[
-\frac{\beta_{s}F}{RT}
\left( \Phi_{1} - R_{\mathrm{SEI}} i_{t} \right)
\right],
\label{eq:sei_current}
\end{equation}
\noindent where $i_{s}$ is the side-reaction current density [A/m$^2$], $k_{f,s}$ the rate constant [m/s], $c_{\mathrm{EC}}$ the solvent concentration in the SEI film [mol/m$^3$], $\Phi_{1}$ the negative-electrode solid-phase potential [V], $R_{\mathrm{SEI}}$ the SEI film resistance [$\Omega\cdot$m$^2$], and $i_t$ the total interfacial current density [A/m$^2$]. Together, Eqs.~\eqref{eq:lam}--\eqref{eq:sei_current} extend the SPMe with capacity-fade pathways capturing both electrode material degradation (LAM) and lithium inventory loss due to side reactions (LLI).

\section{Methodology}



\subsection{Generation and Preprocessing of Synthetic Aging Data}

To emulate the effects of manufacturing variability on battery aging, synthetic datasets were generated for a 30 Ah nickel–manganese–cobalt (NMC) graphite cell using the PyBaMM environment \cite{sulzer2021python}. In particular, cell-to-cell heterogeneity was introduced by varying the active material volume fractions of the positive and negative electrodes, parameters known to strongly influence capacity fade signatures \cite{schindler2021evolution, weng2023differential}. For each batch, applied C-rates vary randomly every 1~kAh throughput within their predefined operating ranges (Table \ref{table:1}). Additionally, each full cycle is simulated at 100\% depth-of-discharge (DoD) and ambient temperature of 298K. 

\begin{table}[h!]
\centering
\caption{C-rate operating sets for each batch}
\label{table:1}
\begin{tabular}{lc}
\hline
\textbf{Batch} & \textbf{C-rate Operating Set} \\
\hline
Batch 1 ($\mathcal{B}_1$) & \{3C, 4C, 5C\} \\
Batch 2 ($\mathcal{B}_2$) & \{1C, 2C, 3C\} \\
Batch 3 ($\mathcal{B}_3$) & \{C/3, C/2, 1C, 2C, 3C, 4C, 5C\} \\
Batch 4 ($\mathcal{B}_4$) & \{C/3, C/2, 1C\} \\
\hline
\end{tabular}
\end{table}

\subsubsection{Source and Target Domain Selection}
\begin{figure}[h!] 
    \begin{center}
       \includegraphics[angle=0,scale=0.4]{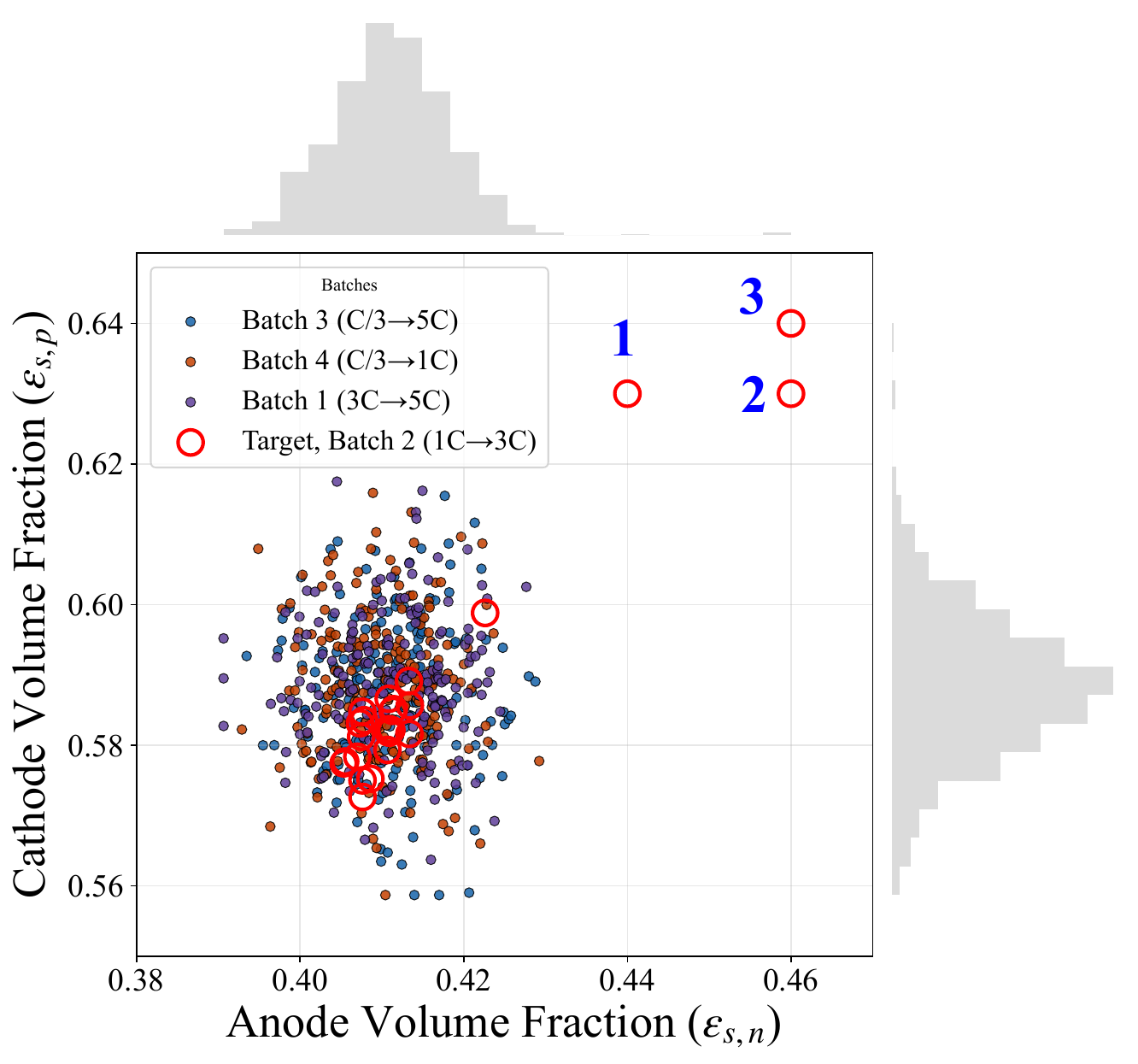}
    \caption{Joint distribution of negative and positive electrode active material volume fractions across batches.}
     \label{fig:2}
    \end{center}
\end{figure}

In batches 1, 3 and 4 (source domain), virtual
cells were created by sampling negative and positive electrode active material volume fractions from a Gaussian distribution centered on nominal values, with standard deviations chosen to span $\pm 5\%$ of the mean. Each sampled pair of $(\varepsilon_{s,n}, \varepsilon_{s,p})$ values corresponds to a distinct synthetic cell, producing unique SOH degradation trajectories. On the other hand, batch 2 (target domain) presents different discharge/charge protocols with $(\varepsilon_{s,n}, \varepsilon_{s,p})$ values in and out of the distribution (Fig. \ref{fig:2}).

\begin{figure}[h!] 
    \begin{center}
       \includegraphics[angle=0,scale=0.4]{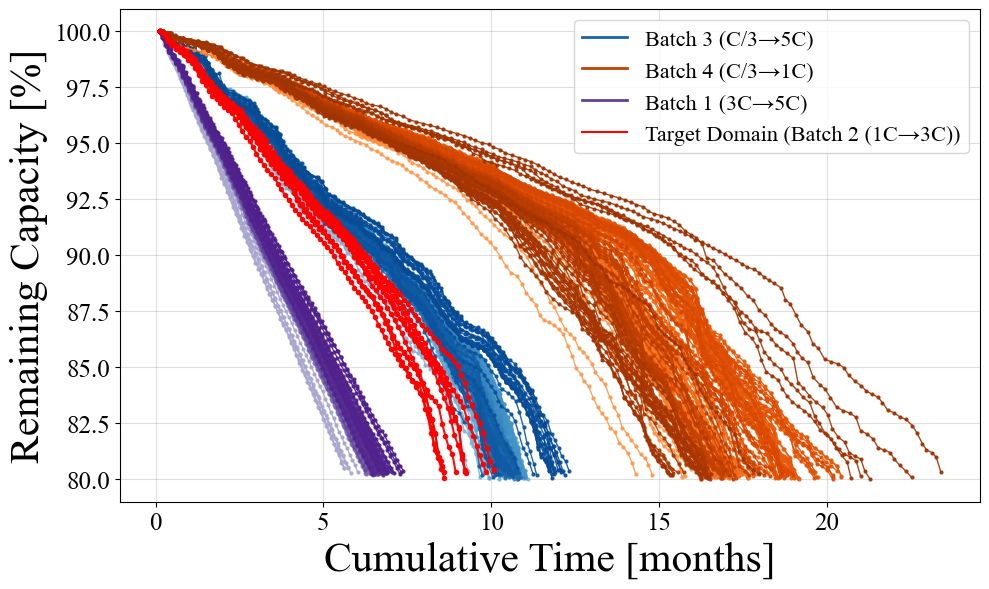}
    \caption{SOH trajectories for each batch.}
     \label{fig:1}
    \end{center}
\end{figure}

Battery cycling and degradation simulations were executed in parallel using a 20-core computing setup, enabling efficient large-scale data generation. The resulting dataset contains diverse degradation behaviors across batches and cell populations, as shown in Figs. \ref{fig:2}-\ref{fig:1}.

\subsubsection{Feature Selection}
In this study, the feature set is constrained to reflect the realistic availability of information in the target domain. Specifically, it is assumed that beyond a cutoff point of 20~kAh throughput, detailed battery behavior such as voltage profiles, differential capacity ($dQ/dV$) shifts, or other electrochemical signatures are not accessible. This assumption reflects practical limitations in real-world scenarios, where only limited diagnostic data may be collected from new or anomalous cells. Consequently, these observables were not considered as input features for reconstructing the SOH trajectories. Instead, only discharge/charge C-rate protocols (the primary features that remain consistently measurable across both source and target domains) will be used for SOH forecasting. 

\subsection{Conformalized Deep Transfer Learning Strategy}

\subsubsection{Deep Learning Architecture Selection} In this study, a Long Short-Term Memory (LSTM) architecture was adopted to address the vanishing gradient issues that typically hinder recurrent neural networks (RNNs) in learning long-term dependencies \cite{kant2025prediction}. LSTMs incorporate gating mechanisms that enable them to capture temporal correlations in battery degradation data, allowing the model to learn complex nonlinear relationships between usage patterns and state of health over extended horizons.

\subsubsection{Transfer Learning Framework}
The proposed transfer learning (TL) framework is illustrated in Fig.~\ref{fig:3}. The source domain $\mathcal{D}_s$ contains a large set of fully labeled cells:
\begin{figure}[h!] 
    \centering
    \includegraphics[angle=0,scale=0.75]{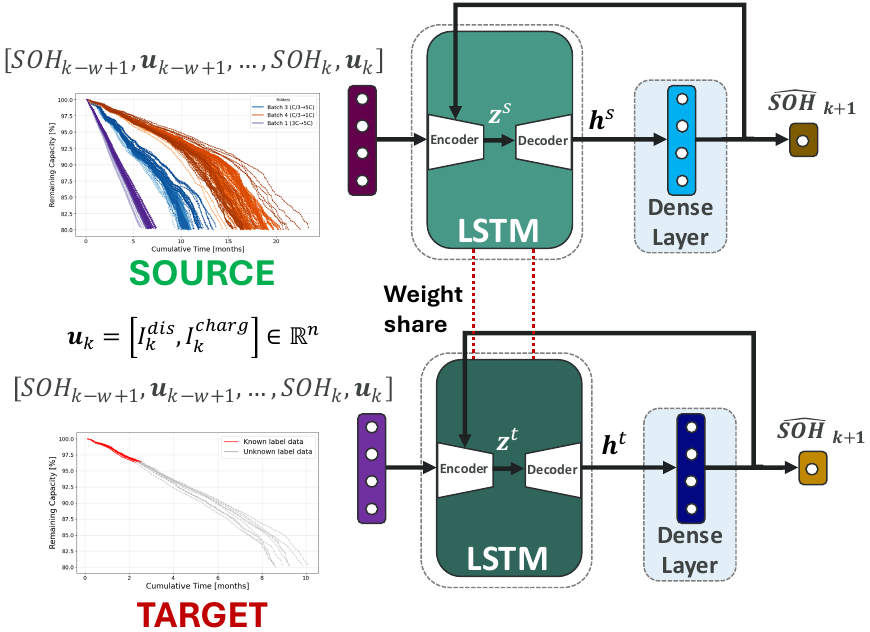}
    \caption{Proposed transfer learning framework combining LSTM-based encoder–decoder with MMD-based domain adaptation.}
    \label{fig:3}
\end{figure}
\begin{equation}
\mathcal{D}_s = \big\{\, \mathbf{x}^{s}, y^{s} \,\big\}
\end{equation}
\noindent whereas the target domain $\mathcal{D}_t$ consists of a limited number of labeled samples (available until the 20~kAh cutoff) and a larger set of unlabeled samples:
\begin{equation}
\mathcal{D}_t = \big\{\, \mathbf{x}^{t}, y^{t} \,\big\} \cup \big\{\, \mathbf{x}^{t} \big\}_{u}
\end{equation}
Each input sample $\mathbf{x}_k$ represents a sequence of $w$ past time steps (sliding window):
\begin{equation}
\mathbf{x}_k = \big[\mathrm{SOH}_{k-w+1}, \mathbf{u}_{k-w+1}, \ldots, \mathrm{SOH}_{k}, \mathbf{u}_{k}\big]
\end{equation}
\noindent with $\mathbf{u}_k = [I_k^{\mathrm{dis}}, I_k^{\mathrm{ch}}]$ denoting the discharge/charge C-rates, and $y_k = \mathrm{SOH}_{k+1}$ being the one-step-ahead label.

\paragraph{Latent representation and decoder output}
The LSTM encoder maps the input sequence into a latent representation:
\begin{equation} \label{Eq7_}
z_k = \mathcal{G}(\mathbf{x}_k; \theta)
\end{equation}
where $\mathcal{G}_{\theta}$ denotes the encoder parameterized by $\theta$. The decoder takes $z_k$ as input and produces the hidden state:
\begin{equation}
h_k = \mathcal{H}(z_k; \psi)
\end{equation}
where $\mathcal{H}$ represents the decoder transformation with parameters $\psi$. The SOH prediction is then obtained via a dense (fully connected) layer:
\begin{equation}
\widehat{\mathrm{SOH}}_{k+1} = \mathcal{F}(h_k; \omega)
\end{equation}
where $\omega$ are the predictor weights.
\paragraph{Domain adaptation.}
To mitigate the distribution shift between source and target domains, Maximum Mean Discrepancy (MMD) is used to align the latent representations $z_k^{s}$ and $z_k^{t}$ in a reproducing kernel Hilbert space (RKHS):
\begin{equation}
\mathrm{MMD}(z^{s},z^{t}) = \left\lVert 
\frac{1}{n_s}\sum_{i=1}^{n_s}\phi(z_i^{s}) - 
\frac{1}{n_t}\sum_{j=1}^{n_t}\phi(z_j^{t})
\right\rVert_{\mathcal{H}}^{2}
\end{equation}

Using the kernel trick, the squared MMD can be expressed as:
\begin{equation}
\begin{split}
\mathrm{MMD}^2(z^{s},z^{t}) &=
\frac{1}{n_s^2}\sum_{i=1}^{n_s}\sum_{j=1}^{n_s} k(z_i^{s},z_j^{s})
+ \frac{1}{n_t^2}\sum_{i=1}^{n_t}\sum_{j=1}^{n_t} k(z_i^{t},z_j^{t})  - \frac{2}{n_s n_t}\sum_{i=1}^{n_s}\sum_{j=1}^{n_t} k(z_i^{s},z_j^{t})
\end{split}
\end{equation}
where $k(\cdot,\cdot)$ is chosen as a Gaussian kernel:
\begin{equation}
k(z_i^{s},z_j^{t}) = \exp\!\left(-\frac{\lVert z_i^{s} - z_j^{t} \rVert^{2}}{2\sigma^{2}}\right)
\end{equation}

\paragraph{Joint optimization}
The network is trained by minimizing a combined loss function:
\begin{equation} \label{eq13_}
\mathcal{L}_{\text{total}}(\theta,\psi,\omega) = 
\underbrace{\sum_{i \in \mathcal{D}_s} 
\big(\mathrm{SOH}_i - \widehat{\mathrm{SOH}}_i\big)^2}_{\mathcal{L}_{\text{source}}(\theta,\psi,\omega)}
+ \lambda\, \underbrace{\mathrm{MMD}\!\big(z^{s},z^{t}\big)}_{\mathcal{L}_{\mathrm{MMD}}(\theta)},
\end{equation}
\noindent where $\lambda$ acts as a regularization coefficient balancing prediction accuracy and domain alignment. To select an appropriate value for $\lambda$ without relying on target-domain labels, a Leave-One-Batch-Out (LOBO) cross-validation procedure is performed entirely within the source domain. In this approach, one batch of source data is withheld during training and used as a pseudo-target domain to evaluate the domain adaptation performance, as described in Algorithm \ref{alg:lobo_lambda}. The hyperparameter $\lambda$ is then chosen to minimize the validation error across all LOBO folds. The optimal model parameters are obtained by solving:
\begin{equation} \label{Eq_14_}
\theta^{*},\, \psi^{*},\, \omega^{*}
= 
\underset{\theta,\psi,\omega}{\arg\min}\;
\mathcal{L}_{\mathrm{total}}(\theta,\psi,\omega)
\end{equation}


\begin{algorithm}
\caption{LOBO for the MMD weight $\lambda$ tuning}\label{alg:lobo_lambda}
\begin{algorithmic}[1]
\State \textbf{Input:} $\Lambda = \{\lambda_1,...,\lambda_K\} \subset [0,1]$, batches $\{\mathcal{B}_1,\ldots,\mathcal{B}_B\}$
\State \textbf{Output:} $\lambda^\star$
\For{$\lambda \in \Lambda$} 
    \State $scores \gets [\,]$
    \For{$b = 1$ \textbf{to} $B$} \Comment{sweeping all source batches}
        \State $\mathcal{D}_{train} \gets \bigcup_{i \neq b}\mathcal{B}_i$
        \Comment{all source batches except $b$}
        \State $\mathcal{D}_{pt} \gets \mathcal{B}_b$ 
        \Comment{pseudo-target batch ($\mathcal{B}_{pt}$)}
        \State Train model parameters $(\theta,\omega,\psi)$ (Eqs. \ref{Eq7_}-\ref{Eq_14_}) \hspace{\algorithmicindent}
        \State $scores \gets scores \cup \{\mathrm{RMSE}(\hat{y}_b, y_b)\}$
        \Comment{RMSE on Batch $b$ ($\mathcal{B}_b$) labels}
    \EndFor
    \State $Score(\lambda) \gets \mathrm{mean}(scores)
    \;=\; \dfrac{1}{B}\sum_{b=1}^{B}\mathrm{RMSE}(\mathcal{B}_b;\lambda)$
\EndFor
\State $\lambda^\star \gets \arg\min_{\lambda \in \Lambda} Score(\lambda)$
\State \Return $\lambda^\star$
\end{algorithmic}
\end{algorithm}

This joint optimization enforces accurate prediction in the source domain while aligning the feature distributions between source and target, enabling improved generalization to anomalous target cells. In practice, this optimization problem is solved using the Adam optimizer, and DL hyperparameters can be found in Table \ref{tab:hyperparams} below.

\begin{table}[h!]
\centering
\caption{Model and training hyperparameters}
\label{tab:hyperparams}
\begin{tabular}{lc}
\hline
\textbf{Parameters} & \textbf{Value} \\
\hline
Sliding window size & 10 \\
Mini-batch size & 32 \\
Hidden size & 256 \\
Learning rate & 10$^{-3}$ \\
\hline
\end{tabular}
\end{table}

To ensure numerical stability during training and to make features comparable in scale, all input variables are standardized. Each feature $x_i$ is standardized as follows:
\begin{equation}
\tilde{x}_i \;=\; \frac{x_i - \mu_x}{\sigma_x}
\end{equation}
where the mean $\mu_x$ and standard deviation $\sigma_x$ are computed from the training dataset (source domain $\mathcal{D}_s$):
\begin{equation}
\mu_x = \frac{1}{|\mathcal{D}_s|}\sum_{j\in\mathcal{D}_s} x_j,~
\sigma_x = \sqrt{\frac{1}{|\mathcal{D}_s|-1}\sum_{j\in\mathcal{D}_s}\!\left(x_j-\mu_x\right)^{2} }
\end{equation}



\subsubsection{Uncertainty Quantification with Conformal Prediction}
Conformal prediction (CP) provides a distribution-free framework for constructing valid prediction intervals for machine learning models, including neural networks, without requiring assumptions about the underlying data distribution and model selection. In this study, uncertainty quantification is performed to forecast the SOH using the domain-adapted LSTM model. To enable statistically valid sequential prediction intervals with temporal dependence, the framework proposed in~\cite{CP_RNN} is adopted. In this approach, conformal prediction intervals are generated for recurrent neural networks (RNNs) by calibrating nonconformity scores over a sliding window. 

To formally define the construction of prediction intervals under this framework, the notion of \emph{nonconformity score} is introduced. These scores quantify the discrepancy between the model's prediction and the true value, and are used to calibrate the uncertainty estimates. Let $N^1, N^2, \ldots, N^n$ be independent random variables with the same distribution, where $N^i$ is considered as the \emph{nonconformity score}. The nonconformity score represents the residual error between the prediction of the underlying model $\hat{y}^i = \Omega(\tilde{y}^i)$ with a given input $\tilde{y}^i$ and the ground truth $y^i$, $N^i = |\hat{y}^i - y^i|$. The goal is to find a prediction region for $N^0$ with a given $N^1, N^2, \ldots, N^n$, particularly the random variable $N^0$ should have a high probability of being included inside the prediction region. For a given significance level or error rate $\alpha \in (0,1)$, we want to build a valid prediction region $\hat{\varepsilon}$. 

According to~\cite[ Lemma 1]{splitCP}, a valid prediction region $\hat{\varepsilon}$ can be considered as $(1-\alpha)$th quantile of the \emph{empirical nonconformity score distribution} $\{N^i\}_{i=1}^n$. By adding $N^{n+1} = \infty$ and assuming that $N^1, N^2, \ldots, N^n$ can be sorted in non-decreasing order, $\hat{\varepsilon}$ can be considered as the $p^{th}$ nonconformity score $N^p$ in which: 
\begin{equation}
p = \left\lceil (n+1)(1-\alpha) \right\rceil
\end{equation}
For a given new example $\tilde{y}^{n+1}$, the conformal prediction region then becomes:
\begin{equation}
\Omega^\alpha(\tilde{y}^{n+1}) = \left[ \hat{y}^{n+1} - \hat{\varepsilon}, \, \hat{y}^{n+1} + \hat{\varepsilon} \right]
\end{equation}
where $\hat{y}^{n+1} = \Omega(\tilde{y}^{n+1})$. This formulation can be adapted to single or multi-step prediction using an LSTM model. Let $\tilde{y}_{t-M-1:t} = (\tilde{y}_{t-M-1}, \tilde{y}_{t-M}, \ldots, \tilde{y}_{t})$ be $M$ historical time-series observations at the time $t$ and the LSTM model can compute predictions $\hat{y}_{t+\tau} = \Omega(\tilde{y}_{t-M-1:t})$ for all future time steps $\tau \in \{1,2,\ldots,H\}$ within the prediction horizon $H$. In this work, derivation of prediction regions is considered for single-step prediction $(H=1)$, where the objective is to ensure that the ground truth SOH value $y_{t+1}$ is contained in the prediction region $\left[\hat{y}_{t+1} - \hat{\varepsilon}, \, \hat{y}_{t+1} + \hat{\varepsilon}\right]$ with a high probability:
\begin{equation}
P\left(y_{t+1} \in \left[ \hat{y}_{t+1} - \hat{\varepsilon}, \, \hat{y}_{t+1} + \hat{\varepsilon}\right]\right)\geq 1 - \alpha
\end{equation}

Let $\Gamma = \{\tilde{y}_{t-M-1:t}, y_t\}_{t=1}^q$ be dataset of time-series observations and split into the training $\Gamma_{\text{train}}$ and calibration $\Gamma_{\text{cal}}$ sets of size $r$ and $q$, respectively. The nonconformity score of each SOH trajectory $y_i \in \Gamma_{\text{cal}}$ can be computed as:
\begin{equation}
N^i = |\hat{y}_i - y_i| \quad \forall~ y_i \in \Gamma_{\text{cal}}
\end{equation}

Assuming that $\{N^i\}_{i=1}^q$ can be sorted in non-decreasing order and adding $N^{q+1}=\infty$, $\hat{\varepsilon}$ can be considered as the $\left\lceil \frac{(q+1)(1-\alpha)}{H} \right\rceil$th smallest nonconformity scores based on the corresponding empirical nonconformity score distribution. For a given set of new observations $\tilde{y}_{t-M-1:t}$, the single-step conformal prediction regions then become:
\begin{equation}
\Omega^\alpha(\tilde{y}_{t-M-1:t}) = \left[\hat{y}_{t+1} - \hat{\varepsilon}, \, \hat{y}_{t+1} + \hat{\varepsilon}\right]
\end{equation}

In this work, 10 cells from each source-domain batch were held out from the training process and instead used to form the calibration dataset. These cells were generated by simulating samples drawn from the same parameter distributions and subjected to the same cycling conditions as the source-domain batches (see Fig.~\ref{fig:1}). This calibration dataset was then employed to compute the empirical nonconformity score distribution. The desired coverage rate was set to 90\% (i.e., $\alpha = 0.1$).

\section{Results and Discussion}
The performance of the proposed framework was first evaluated on the source domain to ensure that the model was capable of accurately learning the underlying degradation dynamics. The LSTM trained purely on the source data achieved agreement with the ground-truth SOH values, yielding an RMSE of $0.083\%$ and an $R^{2}$ score of $0.999$ on the held-out test set (unseen environment). This confirms that the model is able to capture nonlinear dependencies between cycling patterns and SOH evolution when evaluated on the source domain.

\begin{figure}[h!] 
    \begin{center}
       \includegraphics[angle=0,scale=0.50]{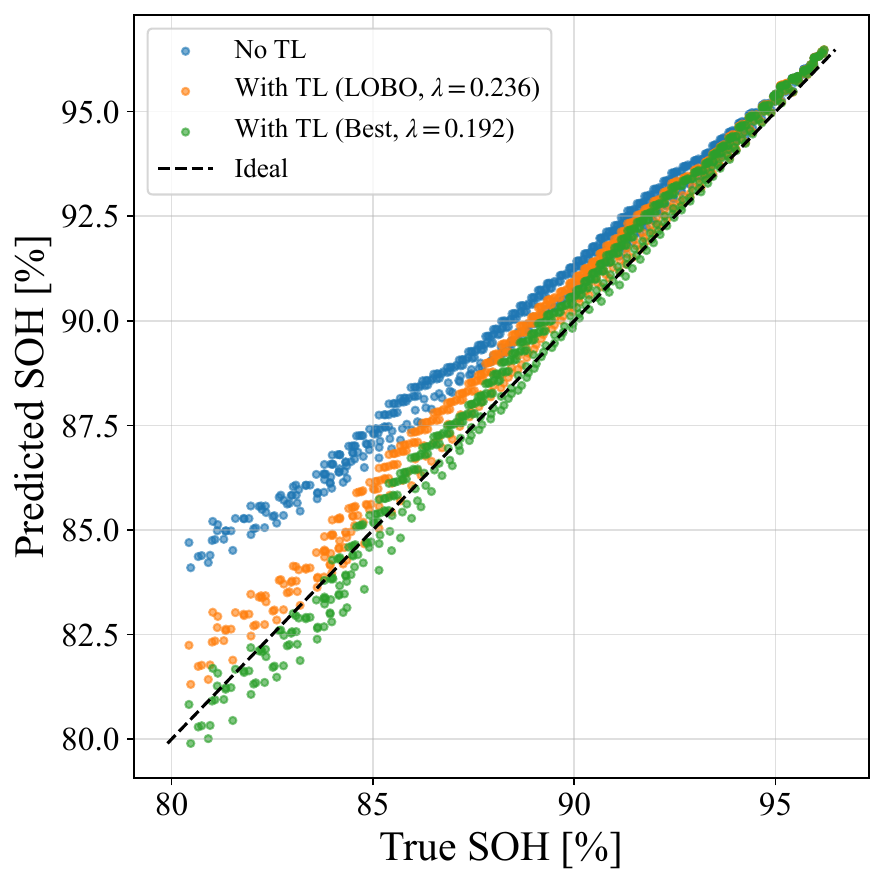}
    \caption{Predictive performance under target domain}
     \label{fig:scatter_plot}
    \end{center}
\end{figure}

However, when directly applied to target cells exhibiting manufacturing-induced variability and different operating conditions, the source-trained model experiences a substantial reduction in accuracy as the battery degradation evolves as shown in Fig. \ref{fig:scatter_plot}, reducing the $R^{2}$ score to 0.833. This domain shift highlights the limitation of purely data-driven models trained on nominal cells, which often fail to extrapolate to unseen domains.

The employment of TL with MMD–based domain adaptation led to substantial improvements in generalization. By aligning the latent feature distributions of the source and target domains, the adapted model achieved an RMSE of $0.781\%$ and $R^{2}=0.962$, corresponding to over $50\%$ reduction in prediction error compared to the baseline (Table \ref{table:II}). Moreover, the domain adaptation weight, $\lambda$, was selected using a Leave-One-Batch-Out (LOBO) cross-validation performed solely on the source data, resulting in $\lambda = 0.236$. This procedure achieved predictive performance on target cells that was only slightly inferior to the ideal case ($\lambda = 0.192$), where the hyperparameter was optimized through grid search with access to target labels. A conventional fine tuning-based TL was also evaluated with dense layer retraining, providing a small improvement over the non-adaptive baseline (error reduction of 2.81\%). This can be explained by the limited supervised target information, which is insufficient to effectively retrain the model to the new domain. By contrast, the proposed method leverages unlabeled target trajectories to enforce domain-invariant latent representations and supports label-free tuning, resulting in improved extrapolation to unseen conditions. 

\begin{table}[h!]
\centering
\caption{Comparison of models under testing in source and target domains}
\label{table:II}
\begin{tabular}{lcc}
\hline
\textbf{Model} & \textbf{RMSE [\%]} & \textbf{R$^2$ Score}\\ 
\hline
LSTM (Source)    &  0.083  & 0.999 \\ 
LSTM (Target)    &  1.637  & 0.833 \\ 
LSTM + TL Fine Tuning (Source)    &  0.084  & 0.999 \\ 
LSTM + TL Fine Tuning (Target)    &  1.591  & 0.843 \\
LSTM + Proposed TL (Source)    &  0.085  & 0.999 \\ 
LSTM + Proposed TL (Target)    &  0.781  & 0.962 \\
\hline
\end{tabular}
\end{table}

\begin{figure*}[!h]
\centering
\subfigure[Target cell 1]{
    \includegraphics[width=0.4515\textwidth]{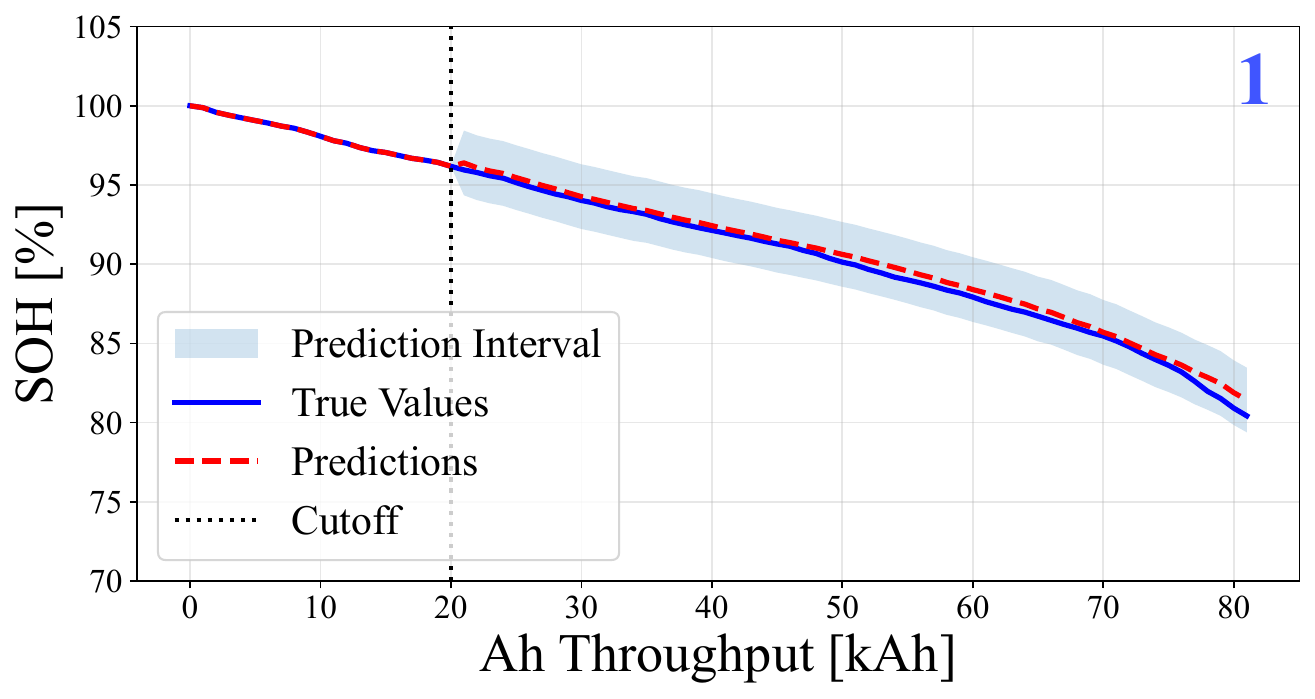}
}
\subfigure[Target cell 2]{
    \includegraphics[width=0.4515\textwidth]{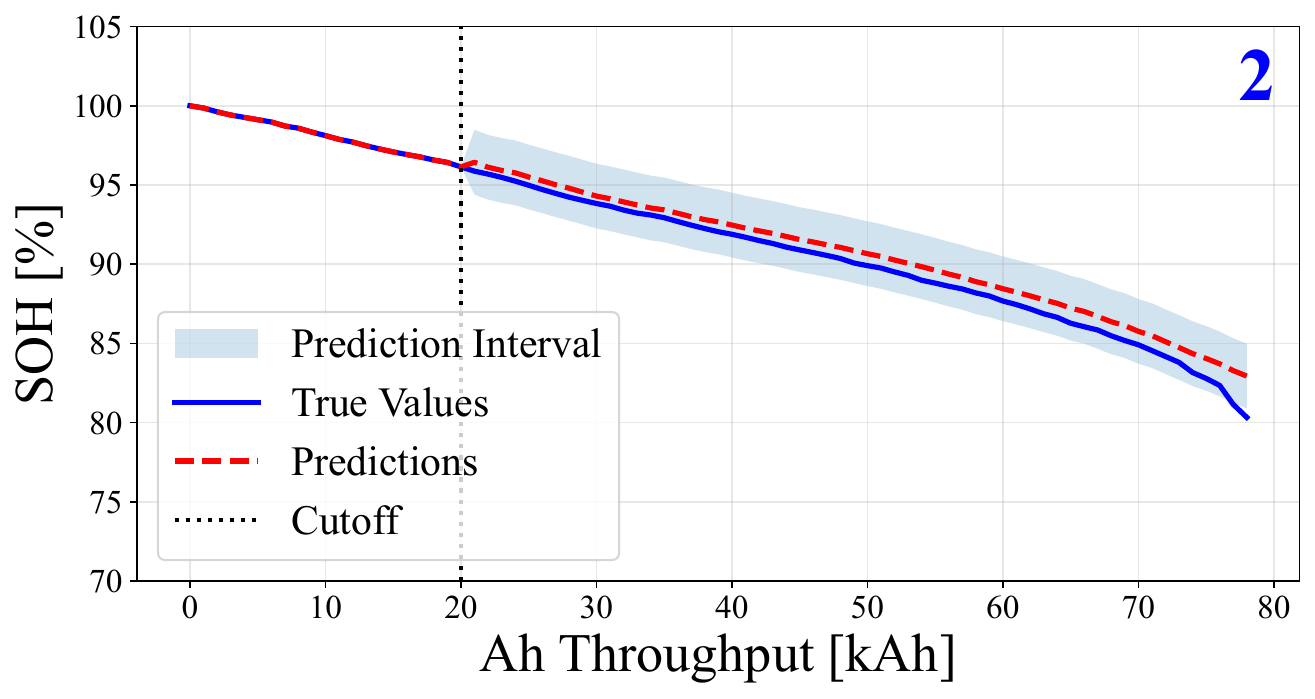}
}
\subfigure[Target cell 3]{
    \includegraphics[width=0.4515\textwidth]{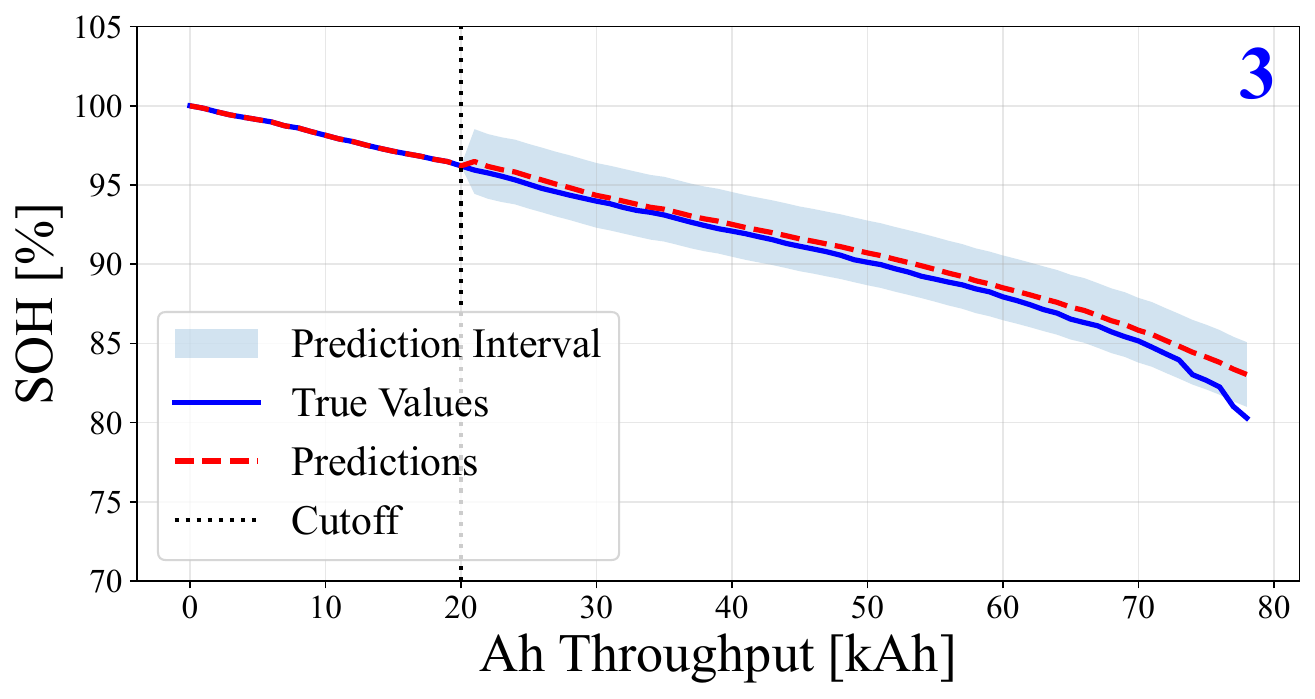}
}
\caption{Conformal prediction for target cells}
\label{fig:PIs}
\end{figure*}

Figure~\ref{fig:PIs} presents the ground-truth SOH trajectories, model predictions, and corresponding prediction intervals (shaded regions) for the three target-domain cells that are most distant from the parameter distribution (see Fig.~\ref{fig:2}). The resulting intervals achieved a nonconformity score of $2.04\%$, corresponding to an average prediction interval width of $4.08\%$ across the prediction horizon. Importantly, although the conformal prediction intervals were calibrated using source-domain cells, the constructed intervals still yielded a mean empirical coverage of $98.8\%$ in the target domain, satisfying the desired coverage rate of $90\%$. This demonstrates that the CP framework not only provides statistically valid uncertainty quantification but also preserves its validity under domain shift, ensuring reliable prediction intervals for SOH forecasting in previously unseen target cells.


\section{Conclusions and Future Work}


This work presents an uncertainty-aware, transfer learning–based framework for robust SOH forecasting under variability induced by manufacturing processes and operating conditions. By combining an LSTM network for temporal modeling, domain adaptation via Maximum Mean Discrepancy (MMD), and Conformal Prediction (CP) for uncertainty quantification, the proposed approach achieves both improved predictive accuracy and calibrated reliability.

The results demonstrate a substantial reduction in prediction error when extrapolating to previously unseen target cells, confirming that aligning latent feature distributions enables effective generalization across heterogeneous cell populations. Moreover, CP provides statistically valid prediction intervals with empirical coverage exceeding the target level, thereby enhancing the trustworthiness of SOH forecasts.

Future work will validate the proposed framework on experimental aging datasets to assess robustness under measurement noise and real-world variability that are not fully captured in synthetic simulations. In particular, benchmarking will be performed using publicly available battery cycling data \cite{luh2024comprehensive}. Additionally, the study will investigate whether hybrid discrepancy-learning strategies for electrochemical models \cite{da2025improving} can be leveraged to improve physics-informed SOH forecasting and generalization under domain shift.

\section*{Acknowledgments}
The authors acknowledge the Honda Research Institute for supporting this work and thank Phillip Aquino for the inspiring discussions and helpful feedback that contributed to the research presented in this paper.

\bibliographystyle{ieeetr}
\bibliography{mybibfile}

\end{document}